\documentclass[letterpaper, 10 pt, conference]{IEEEtran}  %

\usepackage{amsmath} %

\usepackage{footmisc}
\usepackage{tikz}
\usepackage{mathtools}
\usepackage{hyperref}
\usepackage{xcolor}
\usepackage{graphicx}
\definecolor{dark-red}{rgb}{0.4,0.15,0.15}
\definecolor{dark-blue}{rgb}{0.15,0.15,0.8}
\definecolor{medium-blue}{rgb}{0,0,0.5}
\definecolor{medium-teal}{rgb}{0.14,0.56,0.52}
\hypersetup{
    colorlinks, linkcolor={dark-red},
    citecolor={dark-blue}, urlcolor={medium-blue}
}

\newcommand\copyrighttext{%
© © 2022 IEEE. Personal use of this material is permitted. Permission from IEEE must be obtained for all other uses, in any current or future media, including reprinting/republishing this material for advertising or promotional purposes, creating new collective works, for resale or redistribution to servers or lists, or reuse of any copyrighted component of this work in other works.}
\newcommand\copyrightnotice{%
	\begin{tikzpicture}[remember picture,overlay]
		\node[anchor=south,yshift=10pt] at (current page.south) {\fbox{\parbox{\dimexpr\textwidth-\fboxsep-\fboxrule\relax}{\copyrighttext}}};
	\end{tikzpicture}%
}

\IEEEoverridecommandlockouts                              %

\title{%
“If you could see me through my eyes”:\\Predicting Pedestrian Perception %
\thanks{Partially funded by the Federal Ministry of Education and Research (BMBF), project `NEUPA', grant 01IS19078.}
}

\author{\IEEEauthorblockN{Julian Petzold, Mostafa Wahby, Franek Stark, Ulrich Behrje, Heiko Hamann}
\IEEEauthorblockA{\textit{Institute of Computer Engineering} \\
\textit{University of L\"ubeck}\\
L\"ubeck, Germany \\
petzold@iti.uni-luebeck.de}
}

\begin{document}

\maketitle
\thispagestyle{empty}
\pagestyle{empty}

\copyrightnotice
\begin{abstract}
Pedestrians are particularly vulnerable road users in urban traffic. With the arrival of autonomous driving, novel technologies can be developed specifically to protect pedestrians. 
We propose a~machine learning toolchain to train artificial neural networks as models of pedestrian behavior. 
In a~preliminary study, we use synthetic data from simulations of a~specific pedestrian crossing scenario to train a~variational autoencoder and a~long short-term memory network to predict a~pedestrian's future visual perception. We can accurately predict a~pedestrian's future perceptions within relevant time horizons. 
By iteratively feeding these predicted frames into these networks, they can be used as simulations of pedestrians as indicated by our results. 
Such trained networks can later be used to predict pedestrian behaviors even from the perspective of the autonomous car. Another future extension will be to re-train these networks with real-world video data.
\end{abstract}

\begin{IEEEkeywords}
world models, intelligent transportation systems, machine learning, multi-agent systems, sensor prediction
\end{IEEEkeywords}

\section{Introduction}

In recent developments in individual mobility %
we see increasingly automated driver assistance functions. While drivers are relieved by higher degrees of automation, the requirements for functional safety increase. Assistance systems focused on longitudinal vehicle control, such as adaptive cruise control, overtaking assistants, and emergency braking systems, have proven to be advantageous for road safety~\cite{jeong2017evaluating}. However, there are no appropriate substitutes for these radar-based systems in urban areas, where cars share traffic space with vulnerable road users (VRUs)~\cite{yannis2020vulnerable}, such as pedestrians and bicyclists because humans may be difficult to detect by radar~\cite{sun2020mimo}. Vision-based systems  
have the potential to overcome some disadvantages of LIDAR or radar systems. 
They have low costs and increasing capabilities due to advances in artificial intelligence (e.g., deep learning). 
Still, VRUs remain a~major challenge also for camera-based autonomous driving~\cite{sun2019challenges}.
Accident blackspots are found where paths of different road users cross, such as crosswalks and intersections in urban areas~\cite{nguyen2016approach}. 

We propose a~new approach to predict the behavior of VRUs in urban traffic. 
Intuitively, the line of sight (or even eye contact) between a~driver and a~pedestrian standing at the side of the road is important and can provide information about whether a~person is about to enter the road. 
We study how this implicit knowledge can possibly be exploited. 
Using image data, we train artificial neural networks (ANNs) to predict  positions and trajectories of pedestrians in the near future. 
Key to our concept is that we develop agent models of pedestrian behavior using the pedestrian's actual perspective and field of view. 
We believe that with today's technology the pedestrian's perspective can be derived from the data available to the autonomous car. 
A~vehicle's surrounding can already be captured by 360-degree cameras today, and advances in pose recognition~\cite{cao2019openpose} of pedestrians allow to reconstruct their perspective at least partially. 
In this preliminary study, we present our results for the toolchain that works on the pedestrian's view exclusively. 
Initially we focus on pedestrians crossing streets at crosswalks, but our approach is easily extended to address also other VRUs, such as bicyclists and scooter riders. 

We present our results based on synthetic data obtained using the CARLA simulation environment~\cite{Dosovitskiy17}. 
We have developed scenarios that correspond to central European road traffic situations, in particular intersections and crosswalks, as these are the accident blackspots in urban traffic of cars and pedestrians~\cite{nguyen2016approach}. 
We have extended CARLA's semantic segmentation of the simulated camera image to recognize crosswalks as a~separate category. 
For the simulation of pedestrians, we have developed detailed motion sequences and animations including head movements before entering the road (left and right or shoulder check). 
From these pedestrian crossing situations we collect synthetic data and train ANNs, i.e., variational autoencoder (VAE) and long short-term memory (LSTM) network, that can then be used to predict pedestrian behavior. 
For these simulated scenarios, we are able to predict the behavior of a~pedestrian for the immediate future. 
Our main contribution is the toolchain processing pedestrian perspective vision data and outputting a~pedestrian model with predictive power. 
In future work, we plan to apply this toolchain to real-world data and to derive pedestrian perspective images from the autonomous car's 360-degree camera data. 
Our vision is to use these pedestrian models to inform autonomous driving systems about probable dangerous VRU situations with enough look-ahead to initiate a~safe brake actuation. 

\section{Related work}

VRU detection and behavior prediction is an active topic of research for autonomous vehicles~\cite{mannion2019vulnerable}. A~lot of work on VRU detection and behavior prediction focuses on the interaction between vehicles and pedestrians~\cite{rasouli2019autonomous}, but mostly from the perspective of the car~\cite{ahmed2019pedestrian}. Many approaches for pedestrian trajectory prediction are trained and validated on datasets that provide not only images of pedestrians but also action and environmental priors~\cite{kotseruba2016joint, %
robicquet2016learning, rasouli2019pie}. Priors provide additional context %
about tracked agents and environments %
to account for %
feature relevance and improving prediction performance. %

Most %
approaches consider the problem of pedestrian behavior prediction from the perspective of a~vehicle and generally use monocular RGB images as input. Autoencoders are often used to convert data to a~lower-dimensional representation~\cite{bank2020autoencoders}, which helps to process complex data efficiently. %
Ho, Keuper, and Keuper describe another tracking system based on an autoencoder~\cite{ho2020unsupervised}. They use the latent space representation of their visual tracking cues to make their system robust to spatial or temporal changes. Another autoencoder-based technique is variational recurrent neural networks~(VRNN)~\cite{chung2015recurrent}, which include high-level random variables in the latent space of a~%
VAE~\cite{kingma2013auto}. Hoy et al.~\cite{hoy2018learning} use a~VRNN to perform object tracking on the Daimler Pedestrian Path Prediction Dataset~\cite{schneider2013pedestrian}. This tracking is used to generate a~binary crossing/stopping classification of each pedestrian. Poibrenski et al.~\cite{poibrenski2020m2p3, poibrenski2021multimodal} present a~multimodal approach to trajectory prediction that feeds past trajectories and scales of pedestrians into a~conditional autoencoder with RNN decoder-encoder architecture.
Kooij et al.~\cite{kooij2019context} model pedestrian and cyclist trajectories as a~switching linear dynamical system multiple linear models approximate complex nonlinear data dependencies.  %
Using a~stereo camera, their approach switches prediction states based on both, the observed static and dynamic environment as well as %
agent's actions. 
Saadatnejad et al.~\cite{saadatnejad2019pedestrian} deal with generating an invariant, canonical representation of pedestrians in traffic. The authors use OpenPose~\cite{cao2019openpose} and a~generator-discriminator architecture similar to generative adversarial networks (GANs)~\cite{goodfellow2014generative}. A~canonical representation of pedestrians is invariant to different poses, occlusions, transformations and lighting changes.  
It is useful for recognizing and tracking pedestrians that are occluded from the camera for a~long time. %
Rasouli et al.~\cite{rasouli2017they} introduce the use of behavioral and contextual information to improve pedestrian behavior prediction. 
Makansi et al. use mixture density networks to predict the movement and emergence of pedestrians in traffic~\cite{makansi2020multimodal}. They calculate a~reliability prior using semantic segmentation data to determine all possible future positions for a~given object class. Based on this data, the ego-vehicle's motion is compensated, and future pedestrian positions are predicted. 
Mangalam et al. divide the task of pedestrian motion and pose prediction into local and global motion components~\cite{mangalam2020disentangling}. These subproblems are solved using an RNN with a~recurrent encoder-decoder architecture. 
Cao et al.~\cite{cao2020long} predict sequences of human poses in 3D using a~3-stage hierarchical approach. In the first stage, the authors utilize a~single RGB frame and a~history of 2D poses to generate target positions. In the second stage, the paths to these target positions are calculated. Finally, the human poses are estimated.  
Another approach using an encoder-decoder architecture was proposed by Yin et al.~\cite{yin2021multimodal}. Using a~transformer network, they integrate ego-vehicle speed, optical flow and past pedestrian trajectories to predict trajectories. They compensate for the ego-vehicle's motion by %
separate center and pedestrian patches.

Other approaches on pedestrian behavior prediction are not bound to the ego perspective of a~vehicle but use infrastructure-based sensors. %
Zhao et al.~\cite{zhao2019trajectory} present a~pedestrian tracking system that uses LIDAR data fed into a~deep autoencoder neural network. Sun et al.~\cite{sun20183dof} focus on SLAM data. They propose the T-Pose-LSTM, which provides real-time 2D pedestrian trajectory predictions.

Another solution is to use 2D map data %
as training data is readily available~\cite{brvsvcic2013person}. Zhang et al.~\cite{zhang2020prediction} predict if pedestrians jaywalk using a~standard LSTM. They use video data captured by a~camera installed at the crosswalk. Using a~perspective transformation, the authors map pedestrian positions from the video data to a~2D map representation. The LSTM uses location, traffic light state, and several social factors %
to predict if a~given pedestrian will jaywalk. %
Vasquez et al.~\cite{vasquez2014inverse} use Inverse Reinforcement Learning (IRL) to safely navigate a~mobile robot through crowds of pedestrians. Fahad et al.~\cite{fahad2018learning} use IRL as well %
to generate realistic pedestrian trajectories with social interactions on a~2D navigation grid. Employing social affinity maps extracted from human motion trajectories, they train a~deep neural network (DNN), which generates trajectories. The authors focus on the social interactions between pedestrians. Especially the \textit{social forces} technique~\cite{helbing1995social} has proven as a~valid approach for pedestrian trajectory estimation that also takes into account social interactions~\cite{ferrer2013robot}. %
Alahi et al.~\cite{alahi2016social} model the social dependencies in crowds of pedestrians using their Social LSTM, which can model multiple pedestrian trajectories in parallel and thus understand their interactions.
Improving on that, Cheng et al.~\cite{cheng2018pedestrian} use a~Grid LSTM and social pooling to model pedestrian interactions. Chou et al.~\cite{chou2020predicting} use motion features and a~map-based rasterization approach to generate pedestrian trajectories using a~convolutional neural network (CNN) architecture based on MobileNetv2~\cite{sandler2018mobilenetv2}. Ivanovic et al.~\cite{ivanovic2020multimodal} use a~conditional %
VAE to predict pedestrian behavior. Their data-driven approach considers not only the social interactions between pedestrians, but also the movement of vehicles on the road. 

Many of these approaches see pedestrians as black boxes and reduce their interactions with the environment to features that immediately concern their navigation. In this paper, we predict the pedestrian's perspective and consider the pedestrian as an agent with sensory inputs and outputs, providing a~stochastic causal link between the pedestrian's environment and their actions. %
We model such sensory inputs for our prediction model, providing a~world representation for an individual pedestrian.

\section{Methods}

We generate a~realistic pedestrian vision model in the autonomous driving simulator CARLA~\cite{Dosovitskiy17}.
A~considered pedestrian (in the following called ego-pedestrian) 
is controlled by hand-coded state machines to create realistic pedestrian trajectories and head movements. 
We use this pedestrian agent to collect image (view of the pedestrian) and movement data (executed movement actions of the pedestrian) in  simulations. %
For this study, we restrict ourselves to simple street crossing behaviors. The ego-pedestrian approaches a~street crossing, waits if vehicles are approaching, %
and crosses to the opposite sidewalk.
With this data, we train a~VAE~\cite{kingma2013auto} and an LSTM~\cite{hochreiter1997long} (see Secs.~\ref{sec:visualEncDec} and~\ref{sec:predictionModel}). 
We take images from the pedestrian's point of view that are immediately semantically segmented by the simulation environment CARLA. 
The LSTM is provided with input of a~latent vector of a~VAE that encodes the semantically segmented vision. %
First, the collected images are used to train the VAE. Second, the respective latent vectors and sequence of pedestrian actions as movement labels are used to train the LSTM.
Code can be found on our Git repository\footnote{\url{https://gitlab.iti.uni-luebeck.de/petzold/pedestrian_perception}} and supplementary materials on Zenodo~\cite{petzold2021predictingpedestrians}.

\subsection{CARLA and image segmentation}
\label{section:segmentation}

We developed a~pedestrian behavior model for use in virtual 3D traffic environments as in CARLA which is an %
autonomous driving simulator based on Unreal Engine~4.\footnote{\url{https://www.unrealengine.com}}
CARLA uses a~state-of-the-art 3D engine and provides support for detailed pedestrian animations. We created a~small custom map that features a~street that loops in on itself, a~four-way intersection and a~pedestrian crossing. The map is designed to resemble a~central European town, so we use corresponding road signage and right-hand traffic. Using the CARLA API's hierarchical walker skeleton, we control leg, arm, and head joint parameters independently to generate realistic movements. We collect image data from the perspective of the controlled ego-pedestrian. %
We attach a~virtual camera to the ego-pedestrian's head joint, making the camera pose dependent on the ego-pedestrian's  %
pose. %
The head joint is subject to a~chain of 6 kinematic transformations. %
CARLA defines these transformations as 
\emph{base}, \emph{root} and \emph{hips} transformations, followed by two spine transformations (\emph{spine} and \emph{spine01}) and a~\emph{neck} transformation. %

CARLA uses a~fixed time step %
for agent control and %
sensor data collection that we set to 60~ms. 
We collect data using the semantic segmentation camera provided by CARLA, which guarantees an error-free segmentation, as the data is generated based on the built-in object classes. 
To CARLA's 23 object classes, we have added a~class for pedestrian crossings, as they are crucial for our scenario. 
In addition to image data, we collect movement information of the ego-pedestrian %
including flags if the pedestrian has moved this time step, their body's angle parallel to the ground (yaw) in the world coordinate system, and separately their head's rotation parallel to the ground (yaw) in the body's base coordinate system.

\begin{figure}
 \centering
\includegraphics[width=8.5cm,trim={0 0 13.7cm 0},clip]{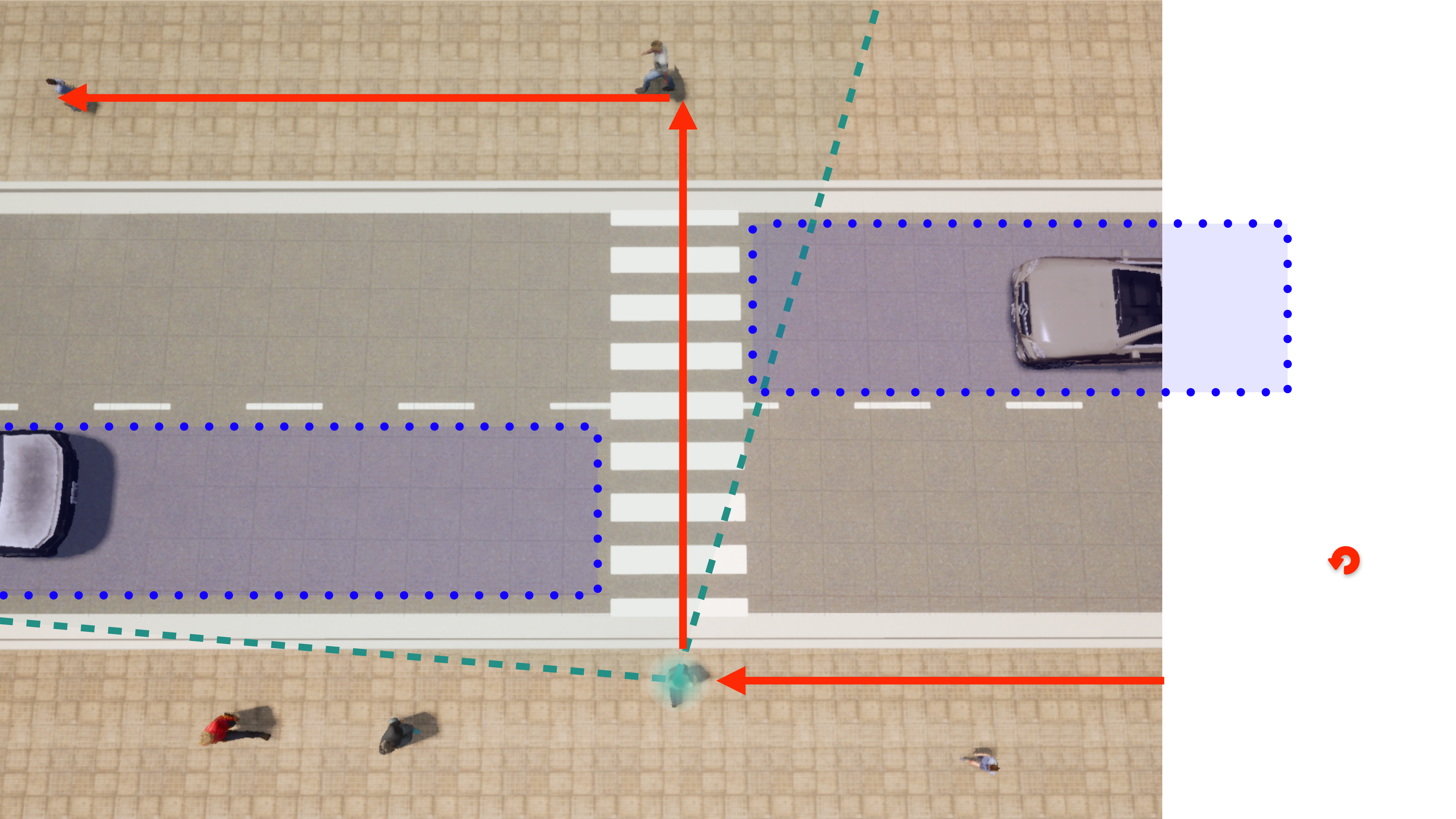}
    \caption{\label{fig:topdown}The crosswalk scenario after 5~s have passed, shown from a~top-down perspective. Several pedestrians approach and leave the crosswalk, as well as cars on both lanes. Directly below the crosswalk the ego-pedestrian (teal) follows the path (red) defined by the FSM in Fig.~\ref{fig:fsm}. Agent detectors placed on the car lanes are marked in blue (dotted boxes). Since the cars are inside at least one detector area, the ego-pedestrian waits in front of the crosswalk in this scenario. The area with green striped borders is the ego-pedestrian's view cone and represents the part of the environment the ego-pedestrian sees.}
\end{figure}

Due to its critical safety, we focus on the ego-pedestrian crossing the street using a~pedestrian crossing. 
For each scenario, we generate random numbers of other pedestrians (uniform distribution unif$\{0, 50\}$, in addition to the ego-pedestrian) and vehicles (unif$\{0, 20\}$). %
The pedestrian and vehicle models are randomly drawn from the blueprint library provided by CARLA. The library provides a~diverse set of models representing adults %
and children for pedestrians and vehicles of different shapes and sizes, including bicycles and motor bikes. Starting locations and destinations
for all pedestrians except the ego-pedestrian are randomly drawn from all traversable sidewalk locations. %
Vehicles spawn from manually placed, but random spawn points on the street. %
Agent movement is controlled by CARLA's simple AI controllers. %

The ego-pedestrian's behavior is implemented using timing-based finite state machines (FSMs). The FSMs realize a~pedestrian trajectory traversing the pedestrian crossing. %
The first FSM controls the ego-pedestrian's two head movements. After $3.6$~s the pedestrian looks to the left for $2.4$~s. After $1.8$~s, the pedestrian looks to the right for $4.2$~s and then looks ahead for the remainder of the scenario.

\begin{figure}
 \centering
\includegraphics[width=8.5cm]{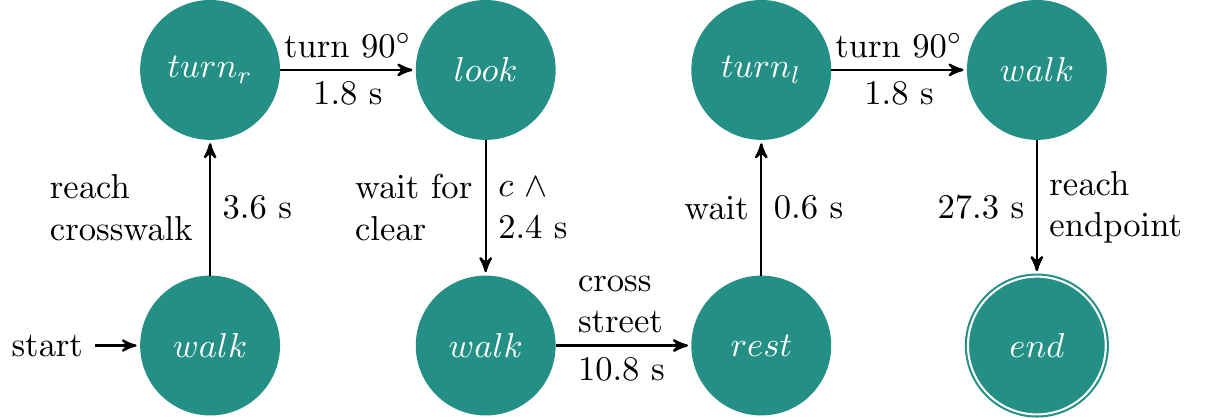}
    \caption{\label{fig:fsm}The ego-pedestrian's behavior, described by a~finite state machine. The walk states are functionally identical and only separated for better readability. Each state transition is labeled with both the waiting time the FSM's transition is based on and a~behavioral description what the ego-pedestrian does during that waiting time. The flag $c$ between the states 
    \emph{look} and \emph{walk} is set when the street in front of the pedestrian crossing is free of approaching cars. The ego-pedestrian's path described by this FSM can be seen in Fig.~\ref{fig:topdown}.}
\end{figure}

The second FSM (see Fig.~\ref{fig:fsm}) manages the movement of the pedestrian's body. 
Initially, the ego-pedestrian is spawned on the sidewalk in the \emph{walk} state. %
Following this, the pedestrian goes through the \emph{turn}$_r$, \emph{look}, \emph{walk}, \emph{rest}, \emph{turn}$_l$ and \emph{walk} states, before finishing the scenario in the \emph{end} state. State transition conditions are mostly based on timing cues chosen based on the ego-pedestrian's walking speed. The only exemption from this is the \emph{wait for clear} transition between \emph{look} and \emph{walk}. It depends on a~timer and on the passing of all cars approaching the crossing inside a~detector area stretching $21$~m along the road for each lane. See Fig.~\ref{fig:fsm} for all details of the second FSM and a~top-down view of our scenario is shown in Fig.~\ref{fig:topdown}.

\subsection{Visual encoder/decoder}
\label{sec:visualEncDec}

We run 1,000 episodes of our simulated pedestrian crossing scenario. At each time step~$t$, we record a~visual representation of the environment~$f_t$ from the pedestrian's perspective, together with its planned action~$a_t$.
This totals in a~dataset $\mathbf{\Psi}$ %
of size $\sim805$k frames that we use to train a~convolutional VAE (ConvVAE) with TensorFlow~\cite{abadi2016tensorflow}. %
The dataset frames were resized to $45\times85$ pixels and modified to be represented in 24~channels (i.e., semantic classes of CARLA plus ours, see Sec.~\ref{section:segmentation}). 
The ConvVAE is required to learn nondeterministic abstract encoded representation~$z$ for each of these input frames, which serve as an input for our prediction model together with action~$a$ (see Fig.~\ref{fig:model}).
This is achieved by sampling from a~probability distribution over all semantic classes for each pixel. %
Note that we allow the overlapping of classes, enabling a~smooth visual transition between segmented objects.
Accordingly, the ConvVAE has a~$45\times85\times24$ input tensor and typical encoder/decoder architecture.
On the one hand, the encoder consists of four chained convolutional layers. The output from these layers is flattened before it is pushed to two separate branching fully connected layers forming the subnetworks, which generate and dense the mean~$\mu$ and logarithmic variance~$\log\sigma^2$ vectors for 50~features.
The~$\mu$ and $\log\sigma^2$ vectors %
are used to sample a~latent vector~$z$.
On the other hand, the decoder is a~transpose of the encoder, and it is trained to decode a~latent vector~$z$ back into the $45\times85\times24$ pixel representation. The output deconvolutional layer uses a~sigmoid function to output normalized pixel values between zero and one, while all other layers use a~Rectified Linear Activation (ReLU) function.
We use the Kullback–Leibler divergence loss function
\begin{equation}
\label{equation:KL_loss}
    \sum_{x \in X} P_{\text{true}}(x)  \log\left(\frac{P_{\text{true}}(x)}{P_{\text{decoded}}(x)}\right)\;
\end{equation}
to evaluate the network's performance by assessing the difference between the original ($P_{\text{true}}$) and decoded ($P_{\text{decoded}}$) distributions. %
We split the dataset~$\mathbf{\Psi}$ into training (90\%) and validation (10\%) and train the ConvVAE for 80 epochs on batches of size~$1,610$ (i.e., two episodes). To prevent overfitting, we pick the network with minimum validation loss at epoch~62.
We use the trained ConvVAE to sort the data~$\mathbf{\Psi}$ by time steps, %
to prepare it for LSTM network training.
Now, a~vector $\mathbf{\psi}_t$ at time step~$t$ is composed of 100 dimensions representing the current encoded frame's~$\mu_t$ and~$\log\sigma_t^2$ for each feature in latent vector~$z_t$, three dimensions %
for ego-pedestrian action~$a_t$, and 100 dimensions 
for~$\mu_{t+1}$ and~$\log\sigma_{t+1}^2$ for each feature at the next time step~$t+1$.

\subsection{Prediction model}
\label{sec:predictionModel}

\begin{figure}
 \centering
\includegraphics[width=8.9cm]{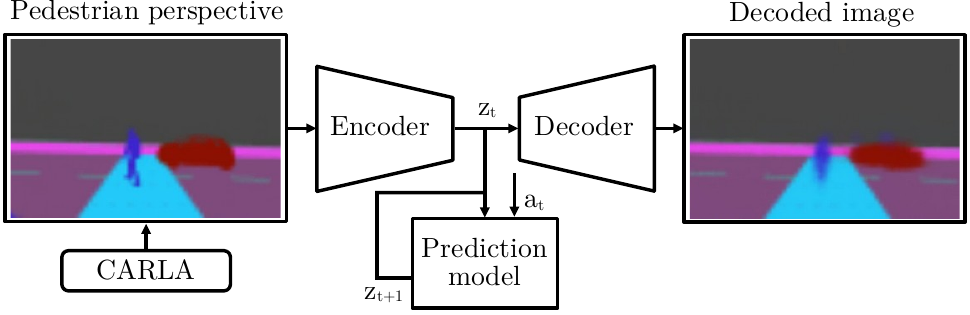}

    \caption{\label{fig:model}Flow diagram showing how, at time step $t$, the VAE encodes the pedestrian perspective of the simulated CARLA agent into a~latent vector $z_t$, which is then input to the prediction model together with the pedestrian's action $a_t$. In turn, the prediction model outputs the next encoded frame $z_{t+1}$, which can be looped back to predict further frames.}
\end{figure}

As a~traffic situation prediction model, we train an LSTM network using TensorFlow and the reformatted dataset~$\mathbf{\Psi}$.
With such a~prediction model, we aim to predict the future pedestrians' surrounding conditions given their current encoded visual perspective and sequence of actions in a~traffic scenario (see Fig.~\ref{fig:model}).
Here, we split %
the data into training (86\%) and validation (14\%) subsets. %
The LSTM network has 53 inputs%
, 512 hidden units, and a~mixture density network~\cite{10.5555/525960} as an output layer (see~\cite{46008} for full architecture information). 
Inspired by Ha et al.~\cite{46008, Ha:2018:RWM:3327144.3327171}, we train the LSTM to output a~probability density function~$p(z)$ %
as a~mixture of Gaussian distribution, instead of deterministic predictions. 
We can control the randomness level by adjusting the temperature parameter~$\tau$ value between zero and one. We obtain deterministic predictions if~$\tau \rightarrow 0$ and increased randomness with higher values.
This enables us to introduce uncertainty and variability, while predicting future surrounding conditions, which is typical in regular traffic situations.
The LSTM receives a~sampled %
encoded frame~$z_t$ and pedestrian action~$a_t$ as input at each time step~$t$ and outputs~$p(z_{t+1})$, 
from which the next encoded frame~$z_{t+1}$ is sampled (see~\cite{46008} for more details).
We train the LSTM in %
batches of size~$1,610$ for $4 \times 10^6$ steps. To prevent overfitting, we use time step $3.8 \times 10^6$ with minimum validation loss.

\begin{figure*}[t]
 \centering
\includegraphics[width=18cm]{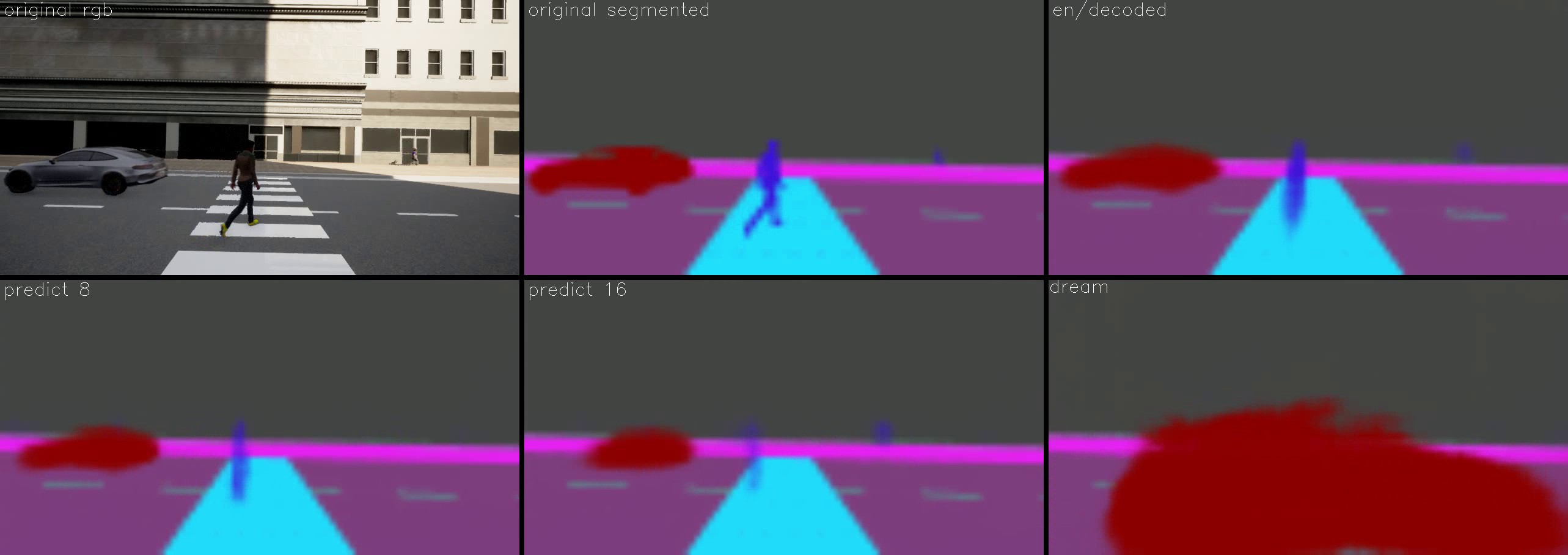}
    \caption{\label{fig:dream}Different modalities of the ego-pedestrian's vision. You see CARLA's RGB representation of our pedestrian's perception (\emph{RGB}); our model's input, the semantically segmented image out of CARLA, at time step $t$ (\emph{original}); the encoded and once again decoded VAE output (\emph{en/decoded}); %
    a~frame decoded from predicted sample $z_{\hat{t}+8}$ with $\hat{t} = t-8$ (\emph{predict 8}); a~frame decoded from predicted sample $z_{\hat{t}+16}$ with $\hat{t} = t-16$ (\emph{predict 16}); 
    our pedestrian's world model (\emph{dream}). The shown predictions for $t$ are based on data captured at previous time steps $\hat{t}$ %
    to allow for direct comparison between all images.
    }
\end{figure*}

\section{Results and Discussion}

Using the VAE and LSTM described in Sections~\ref{sec:predictionModel} and~\ref{sec:visualEncDec}, we have built a~sensory prediction model, and by extension a~world model~\cite{wellmer2021dropout, Ha:2018:RWM:3327144.3327171}, for our pedestrian crossing the street. One prediction time step corresponds to 60~ms, that is the same time horizon as the simulation time step. %
We have evaluated these models in four different experiments, all based on the same scenario set up as described in Sec.~\ref{section:segmentation}. These experiments can be found in our video.\footnote{\url{https://vimeo.com/656754927}\label{fn:repeat}}

\begin{figure}
 \centering
\includegraphics[width=8.5cm,trim={1cm 0.3cm 1cm 1.4cm},clip]{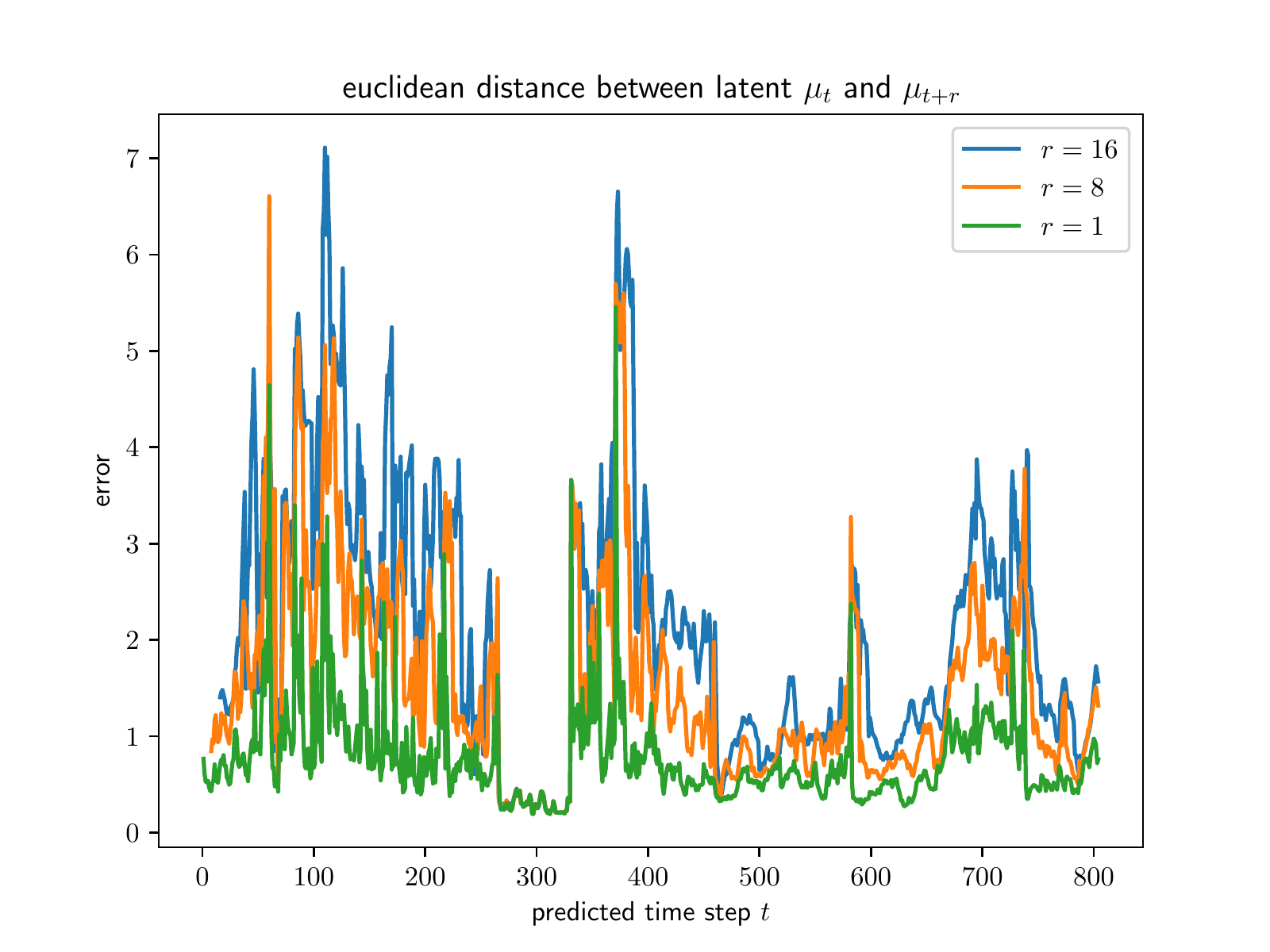}
    \caption{\label{fig:mu_error}Prediction error for one scenario of the pedestrian crossing the street at different prediction horizons $r \in \{1, 8, 16\}$, given as error per time step. The error is the Euclidean distance between $\mu_{t}$ generated by the VAE at $t$ and  prediction $\hat{\mu}^{r}_{t}$ for time step $t$ generated by the LSTM at time step $t-r$.}
\end{figure}

The first experiment is a~quantitative evaluation of the LSTM's capabilities to predict one time step ahead (see error graph Fig.~\ref{fig:mu_error}, %
graph $r=1$).
We generate the one-time-step prediction $\hat{\mu}^{1}_{t}$ (and its corresponding $\log\sigma_{t}^2$) for time step~$t$ based on data from time step $t-1$ by feeding the VAE's encoded frame~$z_{t-1}$ and action~$a_{t-1}$ into our LSTM. The error $e^r_t = ||\hat{\mu}_t^r - \mu_{t}||_2$ for look-ahead $r\in\{1,8,16\}$ is the Euclidean distance between prediction~$\hat{\mu}_t^r$ and the ego-pedestrian's actual recorded data~$\mu_{t}$. $e^r_t$~for $r\in\{8,16\}$ are discussed in the second and third experiment.
As seen in Fig.~\ref{fig:mu_error}, the error graph for $r = 1$ has multiple peaks. There are three kinds of peaks. Some peaks are sudden, without the prediction error increasing significantly before or slowly decreasing after. These high peaks are caused by hard-to-predict movements, often rotations. Large parts of the image are in motion and many pixels %
change classes, resulting in high errors. %
Such errors occurs at $t=60$, 
when the ego-pedestrian turns their body 
to the right 
and their head to the left, or at~$t=85$, when the ego-pedestrian turns their head back to the right (other occurrences: $t\in\{110, 143, 220
\}$). At time steps~$163$ and~$173$ the ego-pedestrian walks while looking to the side, causing some jitter in the movement and peaks in the error graph.
For another type of peak the error rises slowly beforehand. This type is caused by pedestrians or cars appearing in the distance and passing by the ego-pedestrian. Based on the distance and manner of passing, the resulting spike is larger or smaller. This happens at $t = 55$, when a~pedestrian laterally crosses in front of the ego-pedestrian, or at time steps $580, 728$ and $738$, when pedestrians approach frontally. These peaks are rather high, as the pedestrians pass close by the camera. %
At $t = 460$ and $t = 695$, a~pedestrian on the opposite sidewalk and a~car approach. As they are further away, %
their error peaks are lower.

The last kind of error peak appears suddenly and then slowly degrades. This is caused by pedestrians and cars entering the frame directly next to the ego-pedestrian and then moving away from them, e.g., at $t = 265$ and $t=370$. As the sudden appearance of a~car or pedestrian is not predictable, this causes a~high error. 
Time steps~$265$ to~$330$ show a~constant low error without peaks. This happens because the ego-pedestrian does not turn and there are no other traffic participants in their field of view, as they face a~wall.

The second and third experiments evaluate our model's ability to generate valid inputs for larger prediction horizons (Fig.~\ref{fig:mu_error}, %
graphs $r = 8$ and $r = 16$).
In the second experiment we generate prediction $\hat{\mu}^{8}_{t}$ by applying our model 8~times. The first application of our model uses an encoded frame~$z_{t-8}$ and action~$a_{t-8}$ as inputs, further iterations use the previous output frames and corresponding actions. This corresponds to a~prediction 8~time steps ahead and a~prediction horizon of~$\sim 0.5~s$. In the third experiment, the same thing is done 16~times, generating a~prediction 16~time steps ahead and a~prediction horizon of~$\sim 1~s$. Sampled and decoded outputs of these experiments can be seen in Fig.~\ref{fig:dream} and in our video.\footref{fn:repeat}
Both experiments show a~generally higher prediction error than the first experiment %
as they accumulate uncertainty and noise caused by the repeated application of our prediction model (higher errors for~$\hat{\mu}^{16}_{t}$ than~$\hat{\mu}^{8}_{t}$). %
Still, their error plots are qualitatively similar to~$\hat{\mu}^{1}_{t}$. 
Spikes in errors are mostly consistent across all three experiments. 
As the error for~$\hat{\mu}^{1}_{t}$ only depends on data from the previous time step, difficult-to-predict movements of the ego-pedestrian only affect one prediction time step. For~$\hat{\mu}^{8}_{t}$ and~$\hat{\mu}^{16}_{t}$, these errors cause a~`tail' in the error graph, as the prediction model continues to operate on visual input data captured before the sudden movement. Thus, the prediction error is propagated longer for higher prediction horizons.

In our video\footref{fn:repeat}, you can find predictions sampled from $\hat{\mu}^{8}_{t}$ and~$\hat{\mu}^{16}_{t}$ and their corresponding predicted $\log\sigma_{t}^2$. These videos show the stochasticity of the predictions introduced by the sampling process, as the locations of the other pedestrians and cars jump forward and backward along their respective paths. For example, when the pedestrian ahead of the ego-pedestrian exits the crosswalk, %
the 16-time-step prediction (\emph{predict16}) show the pedestrian's position jumping laterally from frame to frame. 
We interpret that as an inherent uncertainty learned by our model (pedestrians may turn left or right at the end of the crosswalk).

Since our pedestrian's world model is capable of predicting the future and given its inherent stochasticity, it can be used to produce new reasonable traffic scenarios of its own.
Here, we present an initial approach to investigating this, given a~predefined sequence of actions to recreate the ego-pedestrian's behavior in the crossing scenario described in Sec.~\ref{section:segmentation}. We loop back the predicted frames into the prediction model (see Fig.~\ref{fig:dream}) and obtain very innovative new crossing scenarios. 
We chose a~relatively high temperature parameter value $\tau = 0.4$, which still outputs scenarios with reasonable randomness. Low temperature values would result in less innovative, more deterministic scenarios, i.e., the world model would devise fewer pedestrians and vehicles into the scene because they are the main cause of randomness.
At $\tau = 0$, the world model doesn't include any pedestrian or vehicle in the output scenarios.
In the example output scenario video\footref{fn:repeat}, although the model receives the same original first image and action sequence from the scenario mentioned above, it produces a~significantly different scenario. For example, the ego-pedestrian finds a~vehicle passing from the left side at the crossing, that doesnt exist in the original scenario (see Fig.~\ref{fig:dream}).

\section{Conclusion}

We have presented our toolchain for (a)~acquiring synthetic data of pedestrians in urban environments, (b)~training a~combination of VAE and LSTM networks, (c)~predicting a~pedestrian's future perception, and (d)~using these networks as simplified simulators of a~pedestrian's environment. %

As we provide only limited information to %
our models during training, the task of predicting other road users' directions of motion may be of increased difficulty. 
Since the LSTM gets only one%
~$z_t$ as an input, it may not be able to detect the direction of movement of pedestrians, if they are only represented as a~blob without movement-defining features. 

By exploiting the inherent stochastic models encoded in the LSTM networks, we can easily generate an ensemble of varied pedestrian behaviors. Similarly, qualitatively different behaviors can be generated by adding more hand-coded finite state machines. 
The scenarios (pedestrian crossing, streets, types and numbers of road users, etc.) can easily be varied. %

We will train %
a simple neural network as a~pedestrian agent exploring our world model. This agent will feed our LSTM with movement commands based on the (predicted) latent vector as input. %
This pedestrian agent's behavior might help to provide more relevant traffic situations to source more training data and improve the world model.

Using this toolchain, we can numerically obtain statistical predictions of future pedestrian positions. This statistical data can then be used to guide the decision making of an autonomous vehicle. 
As future work, we plan to implement the required geometric calculations to derive the considered pedestrian's view based on the autonomous vehicle's perspective. As there will usually be occlusions and hence unknown areas within the pedestrian's visual view, we plan to use machine learning also to realistically fill these gaps~\cite{yu2019free}. 
In future work, we will acquire real-world videos of pedestrian perspectives and expand our approach to translate to the real world.
We are confident that this completed toolchain can be applied in autonomous driving for the advantage of vulnerable road users.

Since the complexity of actively developing traffic scenarios is high, VAE and LSTM might not be suitable modeling tools anymore. Therefore, we will look at transformers~\cite{liu2020convtransformer} to replace both components, encoding and prediction of traffic.

\bibliographystyle{unsrt}
\bibliography{root}

\end{document}